\documentclass[conference]{IEEEtran}
\usepackage{cite}
\usepackage{amsmath,amssymb,amsfonts}
\usepackage{graphicx}
\usepackage{booktabs,multirow}
\usepackage{url}
\usepackage{etoolbox}
\usepackage{listings}
\lstdefinestyle{casecode}{
  language=Verilog,
  basicstyle=\ttfamily\footnotesize,
  columns=fullflexible,
  keepspaces=true,
  showstringspaces=false,
  upquote=true,
  aboveskip=2pt,
  belowskip=2pt
}
\newcommand{\nickname}{CHORUS}

\newcommand{\papermode}{preprint}
\newif\ifreview
\newif\ifcameraready
\ifdefstring{\papermode}{review}{\reviewtrue}{%
  \ifdefstring{\papermode}{preprint}{}{%
    \ifdefstring{\papermode}{camera-ready}{\camerareadytrue}{%
      \PackageError{CHORUS}{Unknown paper mode: \papermode}%
        {Use review, preprint, or camera-ready.}%
    }%
  }%
}

\title{\nickname{}: Coverage-Guided Agentic RL and Policy Consolidation for Testbench Stimulus Generation}
\ifreview
\author{\IEEEauthorblockN{Anonymous Authors}}
\else
\IEEEoverridecommandlockouts
\author{
\IEEEauthorblockN{
Hejia Zhang\textsuperscript{1,*},
Sheng Lu\textsuperscript{2,*},
Zhijing Wu\textsuperscript{1},
Yitong Ma\textsuperscript{1},\\
Zhongming Yu\textsuperscript{1},
Wentao Ni\textsuperscript{1},
Chia-Tung Ho\textsuperscript{3},
Jishen Zhao\textsuperscript{1}}
\IEEEauthorblockA{
\textsuperscript{1}University of California San Diego, La Jolla, CA, USA\\
\textsuperscript{2}Georgia Institute of Technology, Atlanta, GA, USA\\
\textsuperscript{3}NVIDIA, Santa Clara, CA, USA\\
\textsuperscript{1}\texttt{\{hez024,zhw163,yim030,zhy025,w2ni,jzhao\}@ucsd.edu}\\
\textsuperscript{2}\texttt{slu375@gatech.edu}\quad
\textsuperscript{3}\texttt{chiatungh@nvidia.com}}%
\thanks{\textsuperscript{*}Equal contribution.}
}

\fi

\begin{document}
\bstctlcite{chorus:bibcontrol}
\maketitle

\begin{abstract}
Testbenches written by large language models can compile and run while leaving entire datapaths idle.
We present CHORUS, the first framework to combine coverage-guided agentic reinforcement learning (RL) with policy consolidation for testbench stimulus generation.
We train LLM4Cov's final supervised checkpoint to draft testbenches and repair its lowest-scoring draft using simulator feedback, raising CVDP-ECov agentic Pass@1 from 69.2\% to 85.5\%.
Applying the same training to three checkpoints yields similar overall pass rates, yet the policies succeed on different designs.
Averaging their weights combines these strengths in a single dense 4B model that reaches 86.7\% agentic Pass@1.
When simulator rewards select the teacher for each training prompt, the distilled 4B model reaches 88.0\%, 13.5 percentage points above DeepSeek-R1 (671B).
\end{abstract}

\section{Introduction}
\label{sec:intro}
Large language models (LLMs) can translate a design specification into a register-transfer-level (RTL) implementation~\cite{liu2023verilogeval,lu2024rtllm,codevr1}.
To verify that RTL, a stimulus generator must work backward from a behavior in the implementation to an input sequence that triggers it.
Consider a streaming Sobel filter that computes gradients over a window of nine consecutive pixels.
One generated testbench pulses \texttt{valid\_in} for each new pixel.
Each low cycle resets the pixel counter, so the testbench compiles and runs without ever filling the window.

\begin{figure}[t]
\centering
\includegraphics[width=\columnwidth]{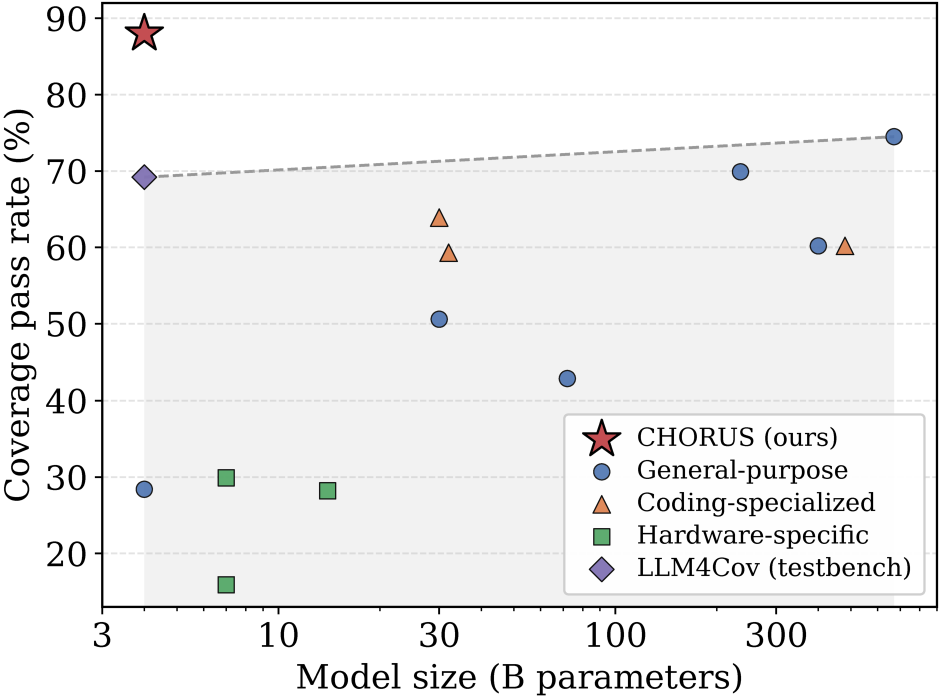}
\caption{CVDP-ECov agentic Pass@1 versus model size (log scale).
\nickname{} (4B) reaches 88.0\% compared with 74.5\% for DeepSeek-R1 (671B).}
\label{fig:motivation}
\end{figure}

The resulting coverage report flags gradient outputs that never toggled.
To repair the testbench, the model must trace those missing toggles back to the input protocol and keep \texttt{valid\_in} high while varying the pixels.
Across the 83 repositories in CVDP-ECov, DeepSeek-R1 (671B) meets the coverage target in 32.5\% of direct-generation trials and 74.5\% when allowed up to three revisions with simulator feedback (Fig.~\ref{fig:motivation} and Table~\ref{tab:overall}).
Feedback more than doubles the pass rate, but roughly one in four trials still ends below target.

Using reinforcement learning (RL) to teach a compact model to act on this feedback poses two challenges.
\textbf{Learning to repair weak drafts.}
When a policy generates several testbenches for the same RTL, its highest-coverage draft may need only a small edit.
A weaker draft gives the model a chance to repair a missing port connection or an input sequence that never reaches the intended state.
\textbf{Retaining stimulus strategies across designs.}
Training three checkpoints from LLM4Cov's supervised fine-tuning (SFT) curriculum~\cite{llm4cov} with the same RL procedure yields policies that finish within 1.2 points of one another on CVDP-ECov.
One reaches the Sobel target in all five trials but omits the module instance in two serial-converter trials.
Another reaches full coverage on the converter in all five trials and misses the Sobel target every time.
Their overall scores give little indication of which policy will cover a particular design.

\begin{figure*}[t]
\centering
\includegraphics[width=\textwidth]{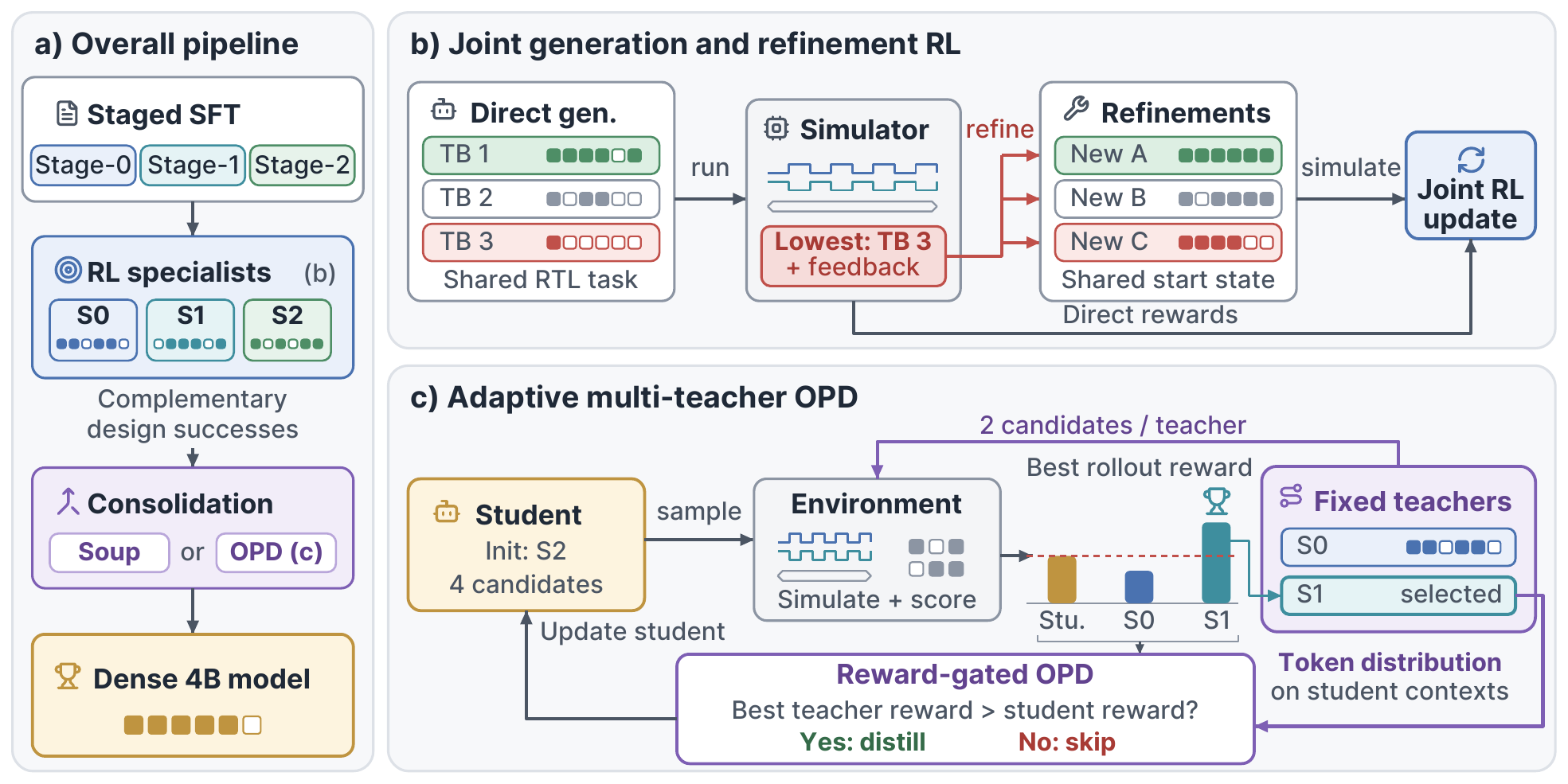}
\caption{\textbf{The \nickname{} pipeline.}
(a) S0, S1, and S2 are the RL specialists trained from the Stage-0, Stage-1, and Stage-2 SFT checkpoints, respectively.
Either consolidation method yields one dense 4B model.
(b) The lowest-coverage direct candidate seeds the entire refinement group; both groups are simulator-scored for one RL update.
(c) S2 initializes the student.
The student and its fixed teachers, S0 and S1, generate candidates for simulator scoring.
The cup marks the best teacher, whose predictions on student contexts are used only when its best reward exceeds the student's; otherwise the prompt is skipped.}
\label{fig:overview}
\end{figure*}

To address these challenges, we propose \nickname{}, a novel post-training framework that uses simulator feedback to teach a model to draft testbenches and repair its weakest drafts.
We train RL specialists from successive SFT checkpoints and consolidate their complementary stimulus strategies into a single dense 4B model for inference (Fig.~\ref{fig:overview}).
\begin{itemize}
\item \textbf{Joint training for drafting and repair.}
We simulate several drafts for each design and generate alternative repairs from the lowest-scoring draft and its simulator report.
Scoring both groups in the same RL update rewards a good first attempt as well as a repair that exercises previously uncovered logic.
Starting from LLM4Cov's final SFT checkpoint, this raises CVDP-ECov Pass@1 from 50.6\% to 74.9\% for the initial draft and from 69.2\% to 85.5\% with feedback.
\item \textbf{Consolidation of complementary RL specialists.}
An oracle choosing a specialist per design reaches 90.8\% agentic Pass@1, compared with 85.5\% for RL on the final SFT checkpoint.
Uniform Model Soup reaches 86.7\% by averaging the three specialists' weights without further training.
Adaptive multi-teacher on-policy distillation (OPD) closes nearly half the gap to the oracle, reaching 88.0\% by distilling from the highest-scoring teacher when it outscores the student on a prompt.
\item \textbf{High-coverage stimuli from a single 4B model.}
Soup gives the highest direct Pass@1 on both AutoEval-ECov and CVDP-ECov, reaching 79.5\% on CVDP-ECov before any repair.
With simulator-guided revisions, OPD exceeds the 671B DeepSeek-R1 by 13.5 points on the same repositories.
In the RGB converter, the distilled model toggles lookup-table data bits that its starting policy left constant.
In the cache, both consolidated models keep the cache enabled between transactions so a later read can hit a dirty line (Fig.~\ref{fig:case-code}).
\end{itemize}

\section{Background and Related Work}
\label{sec:background}
\subsection{LLMs for Hardware Design and Verification}
RTL generation benchmarks~\cite{liu2023verilogeval,lu2024rtllm,fu2026finhardbench,yu2026chipbench} evaluate implementations written by specialized models~\cite{liu2024rtlcoder,codevr1,wei2025vericoder,wang2026verireason} or by agents that invoke design tools~\cite{ho2025verilogcoder,zhao2025mage,yu2026chipmate}.
For verification, AutoBench~\cite{qiu2024autobench} and CorrectBench~\cite{qiu2025correctbench} generate and check testbenches with prompted LLMs.
Repository-level search~\cite{yu2026horizon} uses execution feedback to select among candidate edits at inference time.
PRO-V-R1~\cite{zhao2026prov} trains agents to generate functional reference models and test inputs, while Kochar et al.~\cite{kochar2026testplan} train a test-plan generator with state mutations.

Coverage has long served as an RL reward for directed test generation~\cite{pfeifer2020directed}.
Park et al.~\cite{park2025stimuli} use RL to generate tabular input vectors, while TB or Not TB~\cite{nadimi2025tb} trains on pairs of testbenches ranked offline by simulated coverage.
LLM4Cov~\cite{llm4cov} teaches drafting and repair by imitation, first using a teacher's execution traces and then synthesizing refinements from weak drafts.
CovR~\cite{abdelatty2026covr} uses coverage rewards to train on drafts from the specification and RTL, then deploys the generator in a multi-agent refinement loop.
\nickname{} starts from LLM4Cov's checkpoints and brings the repair into RL by scoring revisions alongside the drafts that prompted them.

\subsection{Policy Optimization, Merging, and Distillation}
A language model's \emph{policy} assigns probabilities to the next token given the RTL, the prompt, and the code written so far.
For a group-relative RL update, the policy samples several candidate testbenches for the same prompt.
DAPO~\cite{dapo} normalizes the rewards within the group and uses these relative scores to favor higher-reward generations.
The policy can therefore learn from the coverage of a completed testbench without differentiating through the simulator.

Once policies have been fine-tuned from a common backbone, their weights can be combined directly.
Uniform Model Soup~\cite{modelsoup} takes the arithmetic mean of corresponding weights, producing one checkpoint without further training.
DARE-TIES~\cite{dare,ties} and DELLA~\cite{della} combine selected parameter changes relative to a base checkpoint to limit interference among updates.
Rewarded Soups~\cite{rame2023rewarded} trains one policy per proxy reward and interpolates their weights to trade off the objectives.
WARP~\cite{rame2024warp} merges policies trained independently from a shared initialization, then interpolates back toward that starting checkpoint to balance reward against policy drift.
In \nickname{}, the specialists start from successive stages of one SFT curriculum.
They train on the same RTL under the same reward, yet meet coverage targets on different designs.

Distillation offers a second way to combine policies by training a student on a teacher's next-token probabilities.
FuseLLM~\cite{wan2024fusellm} trains a single model on next-token distributions from several LLMs evaluated on a shared text corpus.
In \emph{on-policy} distillation, the student generates the code and the teacher scores possible next tokens given the prefix the student actually wrote~\cite{agarwal2024gkd,gu2024minillm}.
For example, the teacher can score continuations of a student-written stimulus loop even if it would have written a different loop itself.
In \nickname{}, the simulator determines which specialist supplies those predictions for each prompt.
The best teacher supervises the student's rollout only when its best testbench outscores the student's best.

\section{The CHORUS Verification Workflow}
\label{sec:method}
\subsection{Joint Cov-Guided Training for Generation and Refinement}
In \emph{direct generation}, the policy drafts a testbench from the RTL, interface specification, and compilation and simulation harness.
In \emph{agentic refinement}, it revises a draft using the simulator diagnostics or coverage report from that attempt (Fig.~\ref{fig:overview}b).
We use DAPO~\cite{dapo} for policy updates, following CodeV-R1~\cite{codevr1}.
CodeV-R1 scores generated RTL with a binary check of output format and functional equivalence to a reference design.
For stimulus generation, a testbench can compile, run, and still exercise only part of the design.
We score each testbench $y$ on design $x$ using measured coverage $c(x,y)\in[0,1]$ and the execution reward
\begin{equation}
R(x,y)=\begin{cases}
1+c(x,y), & \text{valid coverage report},\\
0, & \text{otherwise}.
\end{cases}
\label{eq:reward}
\end{equation}
Compilation failures, simulation failures, and missing or invalid reports receive zero.
Among valid testbenches, each increase in coverage earns a higher reward even below the design's pass threshold.

For each design, the policy first generates a group of testbenches and the simulator scores each one.
We adapt LLM4Cov's worst-state sampling~\cite{llm4cov} to online training by giving the policy its lowest-scoring draft and the corresponding diagnostics or coverage report.
From that same draft the policy samples a group of single-step revisions, which we simulate and score with Eq.~\eqref{eq:reward}.
One group-relative RL update uses both the direct and refinement rollouts.
Section~\ref{sec:diversity} compares the two choices of starting draft.

We apply this procedure independently to the three existing SFT checkpoints, holding the RTL training data, optimizer settings, and update budget fixed, and then compare the resulting policies' aggregate scores and design-level coverage.

\subsection{Consolidating RL Specialists}
\label{sec:opd}
We apply Uniform Model Soup, DARE-TIES, and DELLA to the three RL specialists and evaluate each merged checkpoint directly.
For adaptive OPD, the simulator determines which specialist supplies supervision on each training prompt.

S2 initializes the student, and S0 and S1 serve as fixed teachers during distillation.
For each training task, the student generates four candidates and each teacher generates two, all evaluated by the same simulator reward.
Let $q_s(x)$ and $q_k(x)$ be the best candidate rewards from the student and teacher $k$.
Given these scores, the selection and update rules are
\begin{equation}
\begin{aligned}
k^*(x)&=\arg\max_{k\in\{0,1\}}q_k(x),\\
\operatorname{action}(x)&=\begin{cases}
\text{distill from }k^*, & q_{k^*}(x)>q_s(x),\\
\text{skip}, & \text{otherwise}.
\end{cases}
\end{aligned}
\label{eq:gate}
\end{equation}
Re-evaluating Eq.~\eqref{eq:gate} as the student changes lets teachers supervise different designs without predefined partitions.

On a prompt accepted by Eq.~\eqref{eq:gate}, teacher $\pi_{k^*}$ scores each token $y_t$ sampled by the student $\pi_\theta$ given the same context $c_t$ (the prompt and the rollout so far)~\cite{agarwal2024gkd,gu2024minillm}.
We use Oh et al.'s on-policy distillation with a control variate baseline (vOPD)~\cite{oh2026kl}.
For its reverse-KL update, we compute the token log-ratio $r_t$ and advantage $a_t$ as
\begin{equation}
\begin{aligned}
r_t &= \log \pi_{k^*}(y_t\mid c_t) - \log \pi_\theta(y_t\mid c_t),\\
a_t &= r_t - \hat b_t = r_t + D_{\mathrm{KL}}\!\left(\bar\pi_\theta(\cdot\mid c_t)\,\|\,\bar\pi_{k^*}(\cdot\mid c_t)\right).
\end{aligned}
\label{eq:token-opd}
\end{equation}
The baseline $\hat b_t$ is the negative KL divergence between the barred distributions, each renormalized on the student's $K=16$ most likely tokens at $c_t$.
The log-ratio $r_t$ uses the full-vocabulary probabilities, including when the student samples a token outside this set.
We detach both $r_t$ and the baseline and use student rollouts to estimate the ascent direction
\begin{equation}
\mathbb{E}_{y\sim\pi_\theta}\!\left[\sum_t a_t\,\nabla_\theta\log\pi_\theta(y_t\mid c_t)\right].
\label{eq:opd-gradient}
\end{equation}
At a fixed prefix and teacher, the baseline is independent of the sampled token, so subtracting it leaves the expected token gradient unchanged.

\begin{table*}[!t]
\caption{Pass@1 and mean coverage (Cov@1), in percent.
Bold and underlined entries are the best and second best in each column.
Each CHORUS policy is a dense 4B model.
RL on Best SFT denotes S2, initialized from Stage-2, the strongest supervised checkpoint on CVDP-ECov before RL.
CorrectBench uses a 235B backbone and has no direct-generation result.}
\label{tab:overall}
\centering
\footnotesize
\setlength{\tabcolsep}{3.5pt}
\begin{tabular}{llcccccccc}
\toprule
& & \multicolumn{4}{c}{CVDP-ECov} & \multicolumn{4}{c}{AutoEval-ECov} \\
\cmidrule(lr){3-6}\cmidrule(lr){7-10}
& & \multicolumn{2}{c}{Agentic} & \multicolumn{2}{c}{Direct} & \multicolumn{2}{c}{Agentic} & \multicolumn{2}{c}{Direct} \\
\cmidrule(lr){3-4}\cmidrule(lr){5-6}\cmidrule(lr){7-8}\cmidrule(lr){9-10}
Type & Model & Pass@1 & Cov@1 & Pass@1 & Cov@1 & Pass@1 & Cov@1 & Pass@1 & Cov@1 \\
\midrule
\multirow{5}{*}{\shortstack[l]{General-\\Purpose}}
 & DeepSeek-R1 (671B)        & 74.5\% & 88.5\% & 32.5\% & 51.7\% & \textbf{92.4\%} & \textbf{98.7\%} & 69.4\% & 87.9\% \\
 & Llama-4-Maverick (400B)   & 60.2\% & 81.7\% & 23.1\% & 47.3\% & 32.9\% & 48.7\% & 21.4\% & 41.9\% \\
 & Qwen3-235B-A22B           & 69.9\% & 83.2\% & 29.9\% & 47.0\% & 84.0\% & 92.6\% & 70.6\% & 83.4\% \\
 & Qwen2.5-72B               & 42.9\% & 64.1\% & 16.4\% & 32.2\% & 63.8\% & 87.6\% & 20.9\% & 73.3\% \\
 & Qwen3-30B-A3B             & 50.6\% & 68.0\% & 30.1\% & 52.2\% & 81.9\% & 91.6\% & 72.3\% & 86.1\% \\
\midrule
\multirow{3}{*}{\shortstack[l]{Coding-\\Specialized}}
 & Qwen2.5-Coder-32B         & 59.3\% & 79.6\% & 23.4\% & 52.1\% & 74.5\% & 88.0\% & 52.1\% & 80.1\% \\
 & Qwen3-Coder-30B-A3B       & 63.9\% & 79.9\% & 28.4\% & 48.6\% & 83.8\% & 94.0\% & 75.3\% & 87.1\% \\
 & Qwen2.5-Coder-7B          & 20.5\% & 34.4\% & 11.3\% & 26.6\% & 59.9\% & 77.5\% & 20.9\% & 52.2\% \\
\midrule
\multirow{5}{*}{\shortstack[l]{Hardware /\\Verification}}
 & VeriCoder-Qwen2.5-14B     & 28.2\% & 47.9\% & 17.1\% & 37.3\% & 54.0\% & 69.2\% & 28.3\% & 56.2\% \\
 & CodeV-R1-RL-Qwen-7B       & 29.9\% & 55.5\% & 14.5\% & 32.5\% & 68.5\% & 85.6\% & 42.6\% & 67.9\% \\
 & VeriReason-Qwen2.5-7B     & 15.9\% & 26.4\% &  8.9\% & 16.2\% & 41.5\% & 55.0\% & 13.7\% & 27.9\% \\
 & CorrectBench (235B)       & 34.9\% & 60.5\% &  -- &  -- & 55.8\% & 84.9\% &  -- &  -- \\
 & LLM4Cov-Qwen3-4B          & 69.2\% & 90.4\% & 50.6\% & 76.4\% & 85.0\% & 96.3\% & 79.6\% & 92.6\% \\
\midrule
\multirow{3}{*}{\shortstack[l]{\nickname\\(ours, 4B)}}
 & RL on Best SFT          & 85.5\% & \underline{95.9\%} & 74.9\% & 87.4\% & 88.5\% & 97.0\% & \underline{86.9\%} & \textbf{95.6\%} \\
 & Uniform Soup     & \underline{86.7\%} & 94.9\% & \textbf{79.5\%} & \textbf{89.4\%} & 88.7\% & \underline{97.2\%} & \textbf{87.2\%} & \underline{95.5\%} \\
 & Adaptive OPD     & \textbf{88.0\%} & \textbf{96.0\%} & \underline{78.8\%} & \underline{88.7\%} & \underline{89.0\%} & 97.0\% & 85.1\% & 94.5\% \\
\bottomrule
\end{tabular}
\end{table*}

\section{Experimental Setup}
\label{sec:setup}
\subsection{Benchmarks and Evaluation}
Following LLM4Cov~\cite{llm4cov}, we evaluate on CVDP-ECov, which contains 83 hardware repositories adapted from CVDP~\cite{pinckney2025cvdp} with coverage thresholds set by verification engineers.
AutoEval-ECov contains 156 smaller, single-module tasks derived from VerilogEval~\cite{liu2023verilogeval} using the CorrectBench methodology~\cite{qiu2025correctbench}; its threshold is 100\% coverage.
All simulation and coverage evaluation use Cadence Xcelium \texttt{xrun} 22.03-s001 and IMC 25.09-a001 on Rocky Linux 8.9.

For $M$ tasks and five trials per task, let $c_{ij}$ be coverage, scored as zero for invalid executions, and $\tau_i$ the task threshold.
Pass@1 is $\frac{1}{5M}\sum_{i,j}\mathbf1[c_{ij}\geq\tau_i]$, and Cov@1 is $\frac{1}{5M}\sum_{i,j}c_{ij}$.
Pass@5 is $\frac{1}{M}\sum_i\max_j\mathbf1[c_{ij}\geq\tau_i]$, and Cov@5 is $\frac{1}{M}\sum_i\max_j c_{ij}$.
Agentic Pass@1 on CVDP-ECov is the primary metric.
CovR~\cite{abdelatty2026covr} counts successful execution for Pass@1 and normalizes its coverage score to reference targets, so its published scores use different criteria.
We sample at temperature 0.7 and top-$p$ 0.8; agentic evaluation allows up to three simulator-guided refinements after initial generation.
Our policies use a 16,384-token response cap, while API baselines use their model-specific maximum response lengths.

\subsection{Training and Comparisons}
All \nickname{} policies use the Qwen3-4B backbone initialized from LLM4Cov's SFT checkpoints~\cite{llm4cov}.
We plan to release the consolidated model weights for local inference.
Post-training uses the 11,488-record training split of \texttt{CodeV-R1-11kRTL}, retaining designs with more than 1,000 RTL tokens and exactly one candidate top module.
Training permits two interaction rounds, four rollouts per prompt, and a 360-second EDA timeout.
The three specialists share the same DAPO configuration and learning rate $10^{-6}$.
We sample RL rollouts at temperature 1.0 and use asymmetric clipping ($\varepsilon_{\mathrm{lo}}=0.2$, $\varepsilon_{\mathrm{hi}}=0.28$) with no reference-policy KL penalty ($\beta_{\mathrm{KL}}=0$).
OPD teachers have a 65,536-token context limit.

We report one training run per configuration.
Each specialist trains for 1,000 RL updates (about 100 GPU-hours on two H100 PCIe GPUs), and adaptive OPD runs for 100 further updates on the same training source (about 25 GPU-hours on four H100s).
Table~\ref{tab:overall} compares general-purpose, coding-specialized, and hardware/verification baselines in both evaluation modes.

\section{Results and Analysis}
\label{sec:results}
\subsection{RQ1: How Do RL and Consolidation Improve Coverage?}
Joint RL accounts for the largest gain on the repository-level CVDP-ECov benchmark (Table~\ref{tab:overall}).
Starting from LLM4Cov's strongest supervised checkpoint, it raises agentic Pass@1 from 69.2\% to 85.5\%.
Direct Pass@1 also gains 24.3 points, so more initial drafts already reach their coverage target before the simulator provides feedback.

Consolidation then improves the result under either verification budget.
When only the initial draft counts, Uniform Soup leads on both benchmarks, reaching 79.5\% Pass@1 on CVDP-ECov.
With simulator feedback, DeepSeek-R1 leads on single-module AutoEval-ECov at 92.4\% but loses 17.9 points on the multi-file repositories.
Adaptive OPD stays within one point across the two suites and leads on CVDP-ECov at 88.0\%.

\begin{figure}[t]
\centering
\includegraphics[width=\columnwidth]{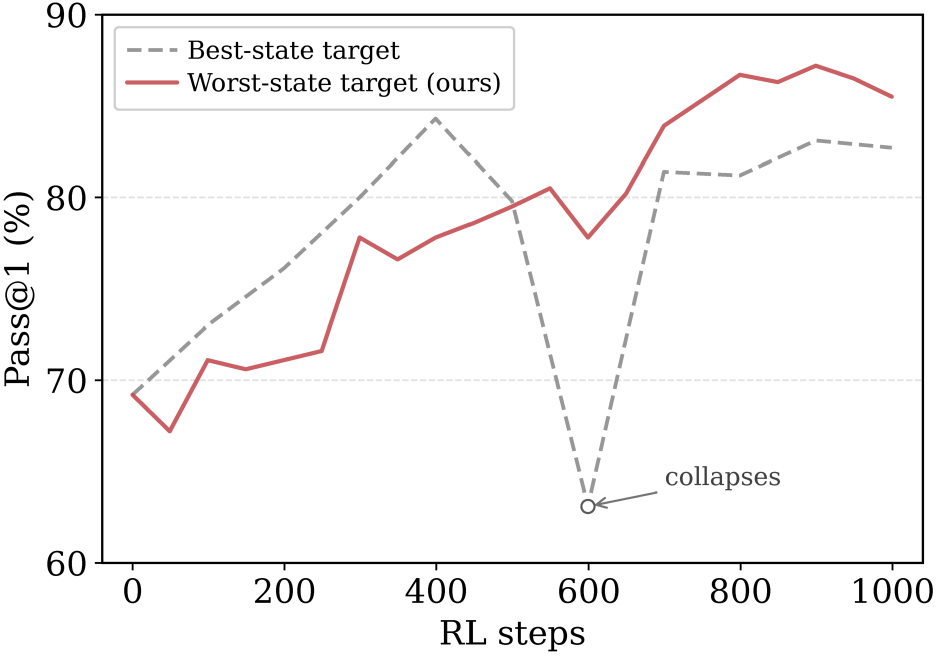}
\caption{Refinement-target ablation from Stage-2 SFT on CVDP-ECov.
Both runs train on drafts and repairs for 1,000 updates, changing which draft seeds each group of repairs.}
\label{fig:refinement}
\end{figure}

\subsection{RQ2: How Does RL Behave Across Repair Targets and Starting Checkpoints?}
\label{sec:diversity}
Which draft should seed the next group of repairs?
Starting from Stage-2, we vary this choice while keeping the joint RL procedure and update budget fixed (Fig.~\ref{fig:refinement}).
A near-complete draft leaves few uncovered bins for a repair to reach.
When the weakest draft fails to compile, every repair sees the build diagnostics, and a revision that produces a valid coverage report outranks one that repeats the failure.
In the strongest-draft run, agentic Pass@1 falls from 84.3\% to 63.1\% between updates 400 and 600, then recovers to 82.7\%.
Training on the lowest-scoring draft avoids that drop and finishes at 85.5\%.

\begin{figure}[t]
\centering
\includegraphics[width=\columnwidth]{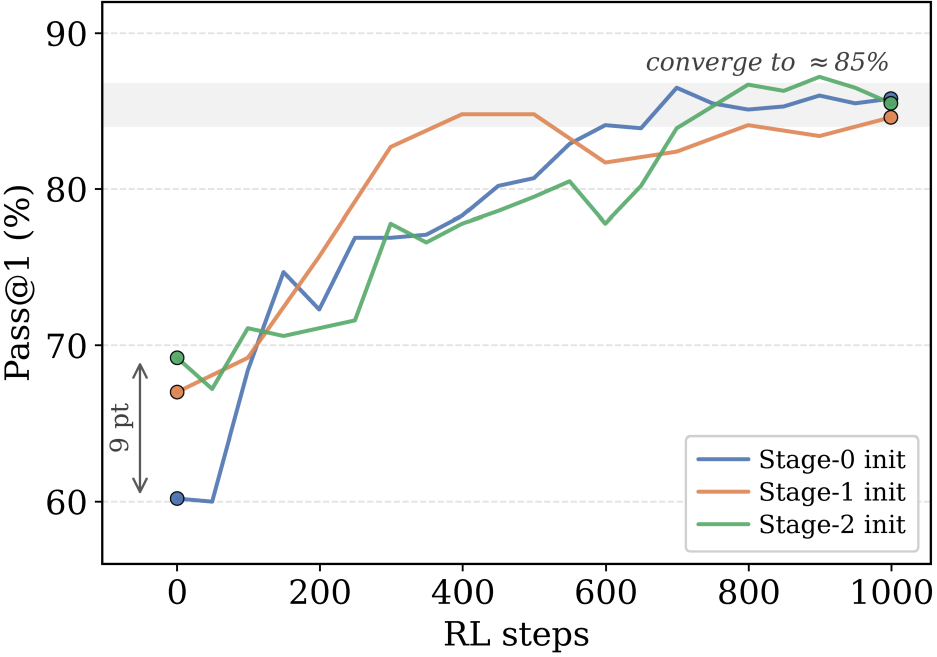}
\caption{CVDP-ECov Pass@1 under three SFT initializations.
The initially weakest checkpoint finishes with the highest Pass@1 after 1,000 updates.}
\label{fig:rl-curves}
\end{figure}

\begin{figure}[t]
\centering
\includegraphics[width=\columnwidth]{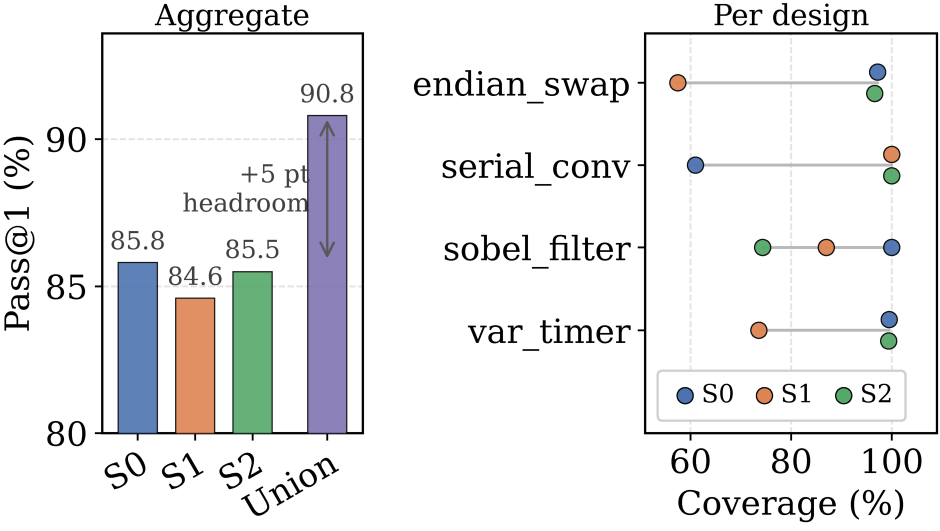}
\caption{Complementarity on CVDP-ECov.
Left, the oracle union exceeds the best RL specialist by five Pass@1 points.
Right, the three specialists (S0, S1, S2) differ in mean coverage on four designs selected for disagreement.}
\label{fig:diversity}
\end{figure}

With the repair rule fixed, does the starting checkpoint still matter?
After 1,000 updates the nine-point spread in supervised Pass@1 has narrowed to 1.2 points (84.6\%--85.8\%), and the initially weakest checkpoint now leads (Fig.~\ref{fig:rl-curves}).
These similar endpoints conceal differences on individual designs.
Choosing the best specialist separately for each repository in hindsight gives 90.8\% agentic Pass@1, five points above the strongest single specialist (Fig.~\ref{fig:diversity}).
Most designs are solved or missed by all three specialists, but which specialist leads changes on the four repositories shown.

On \texttt{serial\_conv}, the stimulus must drive the instantiated converter's three-bit mode port to exercise its eight line codes.
Two of S0's five trials omit the converter instance entirely, while S2 reaches full coverage in every trial.
On the Sobel filter the ranking reverses, with S0 reaching the target in all five trials and S2 missing it every time.
On \texttt{endian\_swap}, an Avalon-ST interface with CSR-controlled byte reversal, partial-coverage reports flag high bits of the 32-bit packet counter and its CSR readback that require longer packet streams.
The oracle uses the observed outcomes to choose a specialist for each design.
Consolidation aims to retain those successes within the inference budget of one policy.

\subsection{RQ3: How Much of the Oracle Gap Can One Model Close?}
\label{sec:ablation}
\textbf{Weight averaging.}
Uniform Soup reaches 86.7\% agentic Pass@1, 0.9 points above the best specialist, with no additional training (Table~\ref{tab:merging}).
Mean coverage falls from the specialists' 95.8\%--95.9\% to 94.9\%, yet small gains near a design's threshold can still turn a miss into a pass.
On the RGB converter, for example, each of Soup's three passing trials clears the 95\% target by just 0.6 points.
Across the three merge bases, DARE-TIES ranges from 77.6\% to 86.0\% Pass@1, and DELLA peaks at 85.8\%.
Averaging leaves 4.1 points between the merged policy and the per-design oracle.

\begin{table}[t]
\caption{RL specialists and static merges on CVDP-ECov under agentic evaluation (\%).
DARE-TIES and DELLA use their best Pass@1 base, S2, at density 0.5.}
\label{tab:merging}
\centering
\footnotesize
\setlength{\tabcolsep}{4pt}
\begin{tabular}{lrrr}
\toprule
Model & Pass@1 & Pass@5 & Cov@1\\
\midrule
S0 (Stage-0 + RL) & 85.8 & 89.2 & 95.8\\
S1 (Stage-1 + RL) & 84.6 & 90.4 & 95.8\\
RL on Best SFT (S2) & 85.5 & 89.2 & 95.9\\
\midrule
Uniform Soup & 86.7 & 91.6 & 94.9\\
DARE-TIES & 86.0 & 92.8 & 95.4\\
DELLA & 85.8 & 90.4 & 96.1\\
\midrule
Oracle union (analysis) & 90.8 & 92.8 & 97.2\\
\bottomrule
\end{tabular}
\end{table}

\textbf{Teacher selection and reward gating.}
Can a student close more of that gap by learning from different teachers on different prompts?
Starting from S2, adaptive OPD reaches 88.0\% after 100 updates, 2.2 points above the strongest individual specialist.
Continuing pure RL from the same student for 100 updates reaches 85.1\% (Table~\ref{tab:ablation}).

The decision is what to do when even the best sampled teacher fails to beat the student.
Always distilling from the better of the two teachers reaches 86.3\% Pass@1.
Using student RL on those unbeaten prompts raises Pass@1 to 87.5\%, and skipping them raises it to 88.0\%.
With skipping fixed, the median-rollout gate reaches 86.7\%.
A single high-coverage sample can therefore qualify a specialist to supervise the student's code under the best-rollout rule.
Both skip rules and unconditional best-teacher distillation reach 91.6\% Pass@5, so their clearest separation is at one trajectory per design.

\begin{table}[t]
\caption{Teacher selection and reward gating on CVDP-ECov after 100 additional updates (\%).
Action gives the update when no teacher beats the student.
Median gates compare median rollout rewards.
Bold scores mark column maxima.}
\label{tab:ablation}
\centering
\footnotesize
\setlength{\tabcolsep}{2.5pt}
\begin{tabular}{llrrrr}
\toprule
Method & Action & Pass@1 & Pass@5 & Cov@1 & Cov@5\\
\midrule
Initial student (S2) & -- & 85.5 & 89.2 & 95.9 & 97.5\\
Continued pure RL & RL & 85.1 & 90.4 & 95.0 & 97.2\\
Always best teacher & Distill & 86.3 & \textbf{91.6} & 95.3 & \textbf{97.7}\\
Reward gate & RL & 87.5 & 90.4 & 95.3 & 97.4\\
\textbf{Adaptive OPD} & Skip & \textbf{88.0} & \textbf{91.6} & \textbf{96.0} & 97.6\\
\midrule
Median gate & RL & 86.0 & 90.4 & 95.8 & 96.8\\
Median gate & Skip & 86.7 & \textbf{91.6} & 95.7 & 96.5\\
\bottomrule
\end{tabular}
\end{table}

Once the three specialists are available, Soup requires only weight averaging, while OPD uses additional teacher rollouts and about 25 GPU-hours of training.

\subsection{RQ4: Which Stimulus Changes Exercise Previously Missed RTL?}
\label{sec:cases}
We trace coverage differences on three repositories to the sequences generated by S2, Soup, and OPD.
Fig.~\ref{fig:case-code} compares initial drafts, and Table~\ref{tab:cases} gives final outcomes over five trials per model and design.

\begin{table}[t]
\caption{Case-study pass rate (Pass) and mean final coverage (Cov.), in percent over five trials.
Coverage targets are in parentheses; S0, S1, and S2 are RL specialists initialized from Stage-0, Stage-1, and Stage-2, respectively.}
\label{tab:cases}
\centering
\footnotesize
\setlength{\tabcolsep}{4.5pt}
\begin{tabular}{lrrrrrr}
\toprule
& \multicolumn{2}{c}{Sobel (100\%)} & \multicolumn{2}{c}{RGB (95\%)} & \multicolumn{2}{c}{Cache (100\%)}\\
\cmidrule(lr){2-3}\cmidrule(lr){4-5}\cmidrule(lr){6-7}
Model & Pass & Cov. & Pass & Cov. & Pass & Cov.\\
\midrule
S0 & 100 & 100.0 & 100 & 95.6 & 100 & 100.0\\
S1 & 60 & 87.0 & 80 & 94.4 & 80 & 98.4\\
S2 & 0 & 74.3 & 0 & 93.8 & 20 & 93.6\\
\midrule
Soup & 100 & 100.0 & 60 & 95.1 & 100 & 100.0\\
OPD & 100 & 100.0 & 80 & 94.7 & 100 & 100.0\\
\bottomrule
\end{tabular}
\end{table}

\begin{figure}[t]
\centering
{\footnotesize\textbf{(a) Sobel: loop bodies, 10 ns clock}\par}
\smallskip
\begin{minipage}[t]{0.48\columnwidth}
{\footnotesize\textbf{S2: pulsed valid}}
\begin{lstlisting}[style=casecode]
pixel_in = 8'd1;
valid_in = 1;
#10;
// Resets pixel counter
valid_in = 0;
#10;
\end{lstlisting}
\end{minipage}\hfill
\begin{minipage}[t]{0.48\columnwidth}
{\footnotesize\textbf{Soup: continuous valid}}
\begin{lstlisting}[style=casecode]
// Vary pixels each cycle
pixel_in = $random % 256;
// Fill the 3x3 window
valid_in = 1;
#10;
\end{lstlisting}
\end{minipage}
\smallskip

{\footnotesize\textbf{(b) RGB conversion: lookup-table write data}\par}
\begin{lstlisting}[style=casecode]
// S2: upper bits stay 0 (i = 0..255)
wdata = 25'h12345 + i;
// S1: sweep all 25 bits (i = 0..99)
wdata = {25{1'b1}} >> i;
// OPD: add random LUT writes (every 100 steps)
wdata = $random % 16777216;
\end{lstlisting}
\smallskip

{\footnotesize\textbf{(c) Cache: loop bodies, 10 ns clock}\par}
\smallskip
\begin{minipage}[t]{0.48\columnwidth}
{\footnotesize\textbf{S2: pulsed enable}}
\begin{lstlisting}[style=casecode]
// Write one cache line
write = 1;
comp = 0;
enable = 1;
#10;
// Clears cache state
enable = 0;
#10;
\end{lstlisting}
\end{minipage}\hfill
\begin{minipage}[t]{0.48\columnwidth}
{\footnotesize\textbf{Soup: persistent state}}
\begin{lstlisting}[style=casecode]
// Keep cache state
enable = 1;
// Mix access/compare ops
comp = $random % 2;
write = $random % 2;
#10;
\end{lstlisting}
\end{minipage}
\caption{Initial-generation excerpts with explanatory comments.
Surrounding code and harness context is partly omitted; statements and timing are retained.}
\label{fig:case-code}
\end{figure}

\textbf{Sobel filter: stimulus timing.}
The streaming $3\times3$ \texttt{sobel\_filter} needs nine consecutive valid pixels to fill its window and variation among those pixels to produce a nonzero gradient.
S2 pulses \texttt{valid\_in} between pixels, resetting the counter each time (Fig.~\ref{fig:case-code}a).
Its testbenches stalled at 67.4\% exercise control signals but miss 30 toggle bins across \texttt{Gx}, \texttt{Gy}, and \texttt{edge\_out}.
Some S1 drafts fill the window with identical pixels by holding each value for ten cycles, while a successful S1 repair changes the pixel each cycle.
Soup and OPD generate varying pixels on consecutive valid cycles even in their initial drafts.
Both reach full coverage in all five trials, as does S0, while S2 misses the target in every trial.

\textbf{RGB conversion: stimulus width.}
The \texttt{rgb\_\allowbreak color\_\allowbreak space\_\allowbreak conversion} module uses a $256\times25$-bit reciprocal lookup table to convert RGB to HSV/HSL.
S2 visits all 256 addresses, but the loop in Fig.~\ref{fig:case-code}b writes values from \texttt{0x12345} to \texttt{0x12444}, using only the low 17 data bits.
Across its trials, \texttt{wdata[24:21]} never toggles, and upper bits of the three reciprocal outputs that feed the color calculations remain uncovered.
S1 sweeps an all-ones mask across the port, while OPD retains student constants and adds randomized writes.
Across five trials, OPD exercises the \texttt{wdata[24]} toggle bin in eight of its 12 generated testbenches, including revisions.
At the 95\% target, OPD passes four of five trials, Soup passes three, and S2 passes none.

\textbf{Direct-mapped cache: transaction state.}
Deasserting \texttt{enable} on a clock edge clears the valid and dirty bits in the 256-line \texttt{direct\_map\_cache}.
A dirty read hit requires an access write, a matching compare write, and a matching compare read with the line intact.
One S2 draft pulses \texttt{enable} between operations, erasing that history (Fig.~\ref{fig:case-code}c).
Another holds \texttt{enable} high but issues only compare writes before clearing the cache.
Soup keeps the cache enabled while mixing access and compare operations, and OPD preserves state across operation loops and revisits selected lines.
Both reach full coverage in all five trials, compared with one for S2.

\section{Conclusion}
\label{sec:conclusion}
\nickname{} consolidates complementary coverage specialists into one dense 4B policy that both drafts testbenches from RTL and revises stimuli when simulation reports missed logic.
Coupling direct generation with weakest-draft revision gives each specialist practice repairing its own coverage gaps across simulator turns.
Carrying intermediate SFT checkpoints through RL is essential because checkpoints with comparable pass rates still discover different working stimuli, such as continuous pixel streams that exercise Sobel gradient logic and transfer cleanly into the consolidated model's initial drafts.

Consolidating these specialists adapts flexibly to verification budgets without serving an ensemble of checkpoints.
When simulation is restricted to a single draft, weight averaging yields the highest direct Pass@1 across both benchmarks with zero additional training; when multi-turn revision is enabled, reward-gated OPD reaches 88.0\% agentic Pass@1 on CVDP-ECov, outperforming DeepSeek-R1 (671B) by 13.5 p.p. while running as one lightweight local 4B model alongside the RTL simulator.
Future work can pair these high-coverage stimuli with specification-derived assertions and reference scoreboards, turning one simulation loop from structural path activation into complete end-to-end functional bug detection.

\clearpage
\label{page:references}
\IEEEtriggeratref{17}
\bibliographystyle{IEEEtran}
\bibliography{references}

@IEEEtranBSTCTL{chorus:bibcontrol,
  CTLuse_forced_etal = {yes},
  CTLmax_names_forced_etal = {6},
  CTLnames_show_etal = {2},
  CTLuse_url = {no}
}

@inproceedings{rame2023rewarded,
  author = {Ram{\'e}, Alexandre and Couairon, Guillaume and Dancette, Corentin and Gaya, Jean-Baptiste and Shukor, Mustafa and Soulier, Laure and Cord, Matthieu},
  title = {Rewarded Soups: Towards {Pareto}-Optimal Alignment by Interpolating Weights Fine-Tuned on Diverse Rewards},
  booktitle = {Proc. NeurIPS},
  year = {2023},
  pages = {71095--71134},
  doi = {10.52202/075280-3114}
}

@misc{rame2024warp,
  author = {Ram{\'e}, Alexandre and Ferret, Johan and Vieillard, Nino and Dadashi, Robert and Hussenot, L{\'e}onard and Cedoz, Pierre-Louis and Sessa, Pier Giuseppe and Girgin, Sertan and Douillard, Arthur and Bachem, Olivier},
  title = {{WARP}: On the Benefits of Weight Averaged Rewarded Policies},
  note = {arXiv:2406.16768},
  year = {2024}
}

@inproceedings{wan2024fusellm,
  author = {Wan, Fanqi and Huang, Xinting and Cai, Deng and Quan, Xiaojun and Bi, Wei and Shi, Shuming},
  title = {Knowledge Fusion of Large Language Models},
  booktitle = {Proc. ICLR},
  year = {2024}
}

@misc{llm4cov,
  author = {Zhang, Hejia and Yu, Zhongming and Ho, Chia-Tung and Ren, Haoxing and Khailany, Brucek and Zhao, Jishen},
  title = {{{LLM4Cov: Execution-Aware Agentic Learning for High-Coverage Testbench Generation}}},
  year = {2026},
  note = {arXiv:2602.16953. ICML 2026},
}

@inproceedings{pfeifer2020directed,
  author = {Pfeifer, N{\'i}colas and Zimpel, Bruno V. and Andrade, Gabriel A. G. and dos Santos, Luiz C. V.},
  title = {A Reinforcement Learning Approach to Directed Test Generation for Shared Memory Verification},
  booktitle = {Proc. DATE},
  year = {2020},
  pages = {538--543}
}

@inproceedings{park2025stimuli,
  author = {Park, Hyeonwoo and Park, Seonghyeon and Kang, Seokhyeong},
  title = {Late Breaking Results: Fine-Tuning {LLMs} for Test Stimuli Generation},
  booktitle = {Proc. DAC},
  year = {2025},
  doi = {10.1109/DAC63849.2025.11132967}
}

@misc{nadimi2025tb,
  author = {Nadimi, Bardia and Filom, Khashayar and Chen, Deming and Zheng, Hao},
  title = {{TB or Not TB}: Coverage-Driven Direct Preference Optimization for {Verilog} Stimulus Generation},
  note = {arXiv:2511.15767},
  year = {2025}
}

@inproceedings{zhao2026prov,
  author = {Zhao, Yujie and Wu, Zhijing and Yuan, Boqin and Yu, Zhongming and Zhang, Hejia and Ni, Wentao and Ho, Chia-Tung and Ren, Haoxing and Zhao, Jishen},
  title = {{PRO-V-R1}: Reasoning Enhanced Programming Agent for {RTL} Verification},
  booktitle = {Proc. DAC},
  year = {2026}
}

@inproceedings{kochar2026testplan,
  author = {Kochar, Dimple Vijay and Pinckney, Nathaniel and Liu, Guan-Ting and Ho, Chia-Tung and Deng, Chenhui and Ren, Haoxing and Khailany, Brucek},
  title = {{GRPO} with State Mutations: Improving {LLM}-Based Hardware Test Plan Generation},
  booktitle = {Proc. ISQED},
  year = {2026}
}

@inproceedings{abdelatty2026covr,
  author = {Abdelatty, Manar and Nouh, Maryam and Reda, Sherief},
  title = {{CovR}: Coverage-Aware Hardware Verification via Reasoning-Guided Reinforcement Learning},
  booktitle = {Proc. ACM/IEEE International Symposium on Machine Learning for CAD (MLCAD)},
  year = {2026},
  doi = {10.1145/3831599.3840347}
}

@inproceedings{fu2026finhardbench,
  author = {Fu, Weimin and Zhang, Hejia and Shao, Minghao and Wang, Zeng and Knechtel, Johann and Sinanoglu, Ozgur and Shafique, Muhammad and Karri, Ramesh and Guo, Xiaolong},
  title = {{FinHardBench}: Can {LLMs} Generate Latency-Aware Hardware for Financial Computing?},
  booktitle = {Proc. COLM},
  year = {2026},
  note = {arXiv:2608.00909}
}

@misc{yu2026chipbench,
  author = {Yu, Zhongkai and Zhou, Chenyang and Lin, Yichen and Zhang, Hejia and Ye, Haotian and Cui, Junxia and Pan, Zaifeng and Zhao, Jishen and Ding, Yufei},
  title = {{ChipBench}: A Next-Step Benchmark for Evaluating {LLM} Performance in {AI}-Aided Chip Design},
  note = {arXiv:2601.21448},
  year = {2026}
}

@misc{yu2026chipmate,
  author = {Yu, Zhongkai and Lin, Yichen and Zhou, Chenyang and Zhang, Yuwei and Zhou, Kun and Cui, Junxia and Ye, Haotian and Hu, Zhengding and Pan, Zaifeng and Wang, Ruiyi and Zhao, Yujie and Zhang, Hejia and Shang, Jingbo and Zhao, Jishen and Ding, Yufei},
  title = {{ChipMATE}: Multi-Agent Training via Reinforcement Learning for Enhanced {RTL} Generation},
  note = {arXiv:2605.12857},
  year = {2026}
}

@inproceedings{liu2023verilogeval,
  author = {Liu, Mingjie and Pinckney, Nathaniel and Khailany, Brucek and Ren, Haoxing},
  title = {{Invited Paper: VerilogEval: Evaluating Large Language Models for Verilog Code Generation}},
  booktitle = {Proc. IEEE/ACM ICCAD},
  volume = {},
  number = {},
  pages = {1-8},
  year = {2023},
  doi = {10.1109/ICCAD57390.2023.10323812},
}

@inproceedings{lu2024rtllm,
  author = {Lu, Yao and Liu, Shang and Zhang, Qijun and Xie, Zhiyao},
  title = {{RTLLM: An Open-Source Benchmark for Design RTL Generation with Large Language Model}},
  booktitle = {Proc. ASP-DAC},
  volume = {},
  number = {},
  pages = {722-727},
  year = {2024},
  doi = {10.1109/ASP-DAC58780.2024.10473904},
}

@inproceedings{liu2024rtlcoder,
  author = {Liu, Shang and Fang, Wenji and Lu, Yao and Zhang, Qijun and Zhang, Hongce and Xie, Zhiyao},
  title = {{RTLCoder: Outperforming GPT-3.5 in Design RTL Generation with Our Open-Source Dataset and Lightweight Solution}},
  booktitle = {Proc. IEEE LLM Aided Design Workshop (LAD)},
  volume = {},
  number = {},
  pages = {1-5},
  year = {2024},
  doi = {10.1109/LAD62341.2024.10691788},
}

@inproceedings{codevr1,
  author = {Zhu, Yaoyu and Huang, Di and Lyu, Hanqi and Zhang, Xiaoyun and Li, Chongxiao and Shi, Wenxuan and Wu, Yutong and Mu, Jianan and Wang, Jinghua and zhao, Yang and Jin, Pengwei and Cheng, Shuyao and Liang, shengwen and zhang, xishan and Zhang, Rui and Du, Zidong and Guo, Qi and Hu, Xing and Chen, Yunji},
  title = {{QiMeng-CodeV-R1: Reasoning-Enhanced Verilog Generation}},
  booktitle = {Advances in Neural Information Processing Systems},
  volume = {38},
  pages = {154266--154300},
  year = {2025},
}

@inproceedings{wei2025vericoder,
  author = {Anjiang Wei and Huanmi Tan and Tarun Suresh and Daniel Mendoza and Thiago S. F. X. Teixeira and Ke Wang and Caroline Trippel and Alex Aiken},
  title = {{VeriCoder: Enhancing {LLM}-Based {RTL} Code Generation through Functional Correctness Validation}},
  booktitle = {NeurIPS Workshop on Deep Learning for Code},
  year = {2025},
  note = {{OpenReview}: aAOStQGcT9},
}

@inproceedings{wang2026verireason,
  author = {Yiting Wang and Guoheng Sun and Wanghao Ye and Gang Qu and Ang Li},
  title = {{VeriReason: Reinforcement Learning with Testbench Feedback for Reasoning-Enhanced Verilog Generation}},
  booktitle = {Proc. ACM GLSVLSI},
  year = {2026},
  pages = {948--953},
  doi = {10.1145/3787109.3815310},
}

@inproceedings{ho2025verilogcoder,
  author = {Ho, Chia-Tung and Ren, Haoxing and Khailany, Brucek},
  title = {{VerilogCoder}: Autonomous {Verilog} coding agents with graph-based planning and abstract syntax tree ({AST})-based waveform tracing tool},
  booktitle = {Proc. AAAI},
  volume = {39},
  pages = {300--307},
  year = {2025},
}

@inproceedings{zhao2025mage,
  author = {Zhao, Yujie and Zhang, Hejia and Huang, Hanxian and Yu, Zhongming and Zhao, Jishen},
  title = {{MAGE: A Multi-Agent Engine for Automated RTL Code Generation}},
  booktitle = {Proc. ACM/IEEE DAC},
  volume = {},
  number = {},
  pages = {1-7},
  year = {2025},
  doi = {10.1109/DAC63849.2025.11133191},
}

@inproceedings{qiu2024autobench,
  author = {Qiu, Ruidi and Zhang, Grace Li and Drechsler, Rolf and Schlichtmann, Ulf and Li, Bing},
  title = {{AutoBench: Automatic Testbench Generation and Evaluation Using LLMs for HDL Design}},
  booktitle = {Proc. ACM/IEEE MLCAD},
  year = {2024},
  doi = {10.1145/3670474.3685956},
}

@inproceedings{qiu2025correctbench,
  author = {Qiu, Ruidi and Zhang, Grace Li and Drechsler, Rolf and Schlichtmann, Ulf and Li, Bing},
  title = {{CorrectBench: Automatic Testbench Generation with Functional Self-Correction using LLMs for HDL Design}},
  booktitle = {Proc. DATE},
  volume = {},
  number = {},
  pages = {1-7},
  year = {2025},
  doi = {10.23919/DATE64628.2025.10992873},
}

@misc{yu2026horizon,
  author = {Cunxi Yu and Chenhui Deng and Nathaniel Pinckney and Brucek Khailany},
  title = {{Agentic Hardware Design as Repository-Level Code Evolution}},
  year = {2026},
  note = {arXiv:2606.28279},
}

@inproceedings{dapo,
  author = {Yu, Qiying and Zhang, Zheng and Zhu, Ruofei and Yuan, Yufeng and Zuo, Xiaochen and Yue, Yu and Dai, Weinan and Fan, Tiantian and Liu, Gaohong and liu, juncai and Liu, LingJun and Liu, Xin and Lin, Haibin and Lin, Zhiqi and Ma, Bole and Sheng, Guangming and Tong, Yuxuan and Zhang, Chi and Zhang, Mofan and Zhang, Ru and Zhang, Wang and Zhu, Hang and Zhu, Jinhua and Chen, Jiaze and Chen, Jiangjie and Wang, Chengyi and Yu, Hongli and Song, Yuxuan and Wei, Xiangpeng and Zhou, Hao and Liu, Jingjing and Ma, Wei-Ying and Zhang, Ya-Qin and Yan, Lin and Wu, Yonghui and Wang, Mingxuan},
  title = {{DAPO: An Open-Source LLM Reinforcement Learning System at Scale}},
  booktitle = {Advances in Neural Information Processing Systems},
  volume = {38},
  pages = {113222--113244},
  year = {2025},
}

@inproceedings{modelsoup,
  author = {Wortsman, Mitchell and Ilharco, Gabriel and Gadre, Samir Ya and Roelofs, Rebecca and Gontijo-Lopes, Raphael and Morcos, Ari S and Namkoong, Hongseok and Farhadi, Ali and Carmon, Yair and Kornblith, Simon and Schmidt, Ludwig},
  title = {{Model soups: averaging weights of multiple fine-tuned models improves accuracy without increasing inference time}},
  booktitle = {Proc. ICML},
  volume = {162},
  pages = {23965--23998},
  year = {2022},
}

@inproceedings{ties,
  author = {Yadav, Prateek and Tam, Derek and Choshen, Leshem and Raffel, Colin and Bansal, Mohit},
  title = {{TIES-Merging: Resolving Interference When Merging Models}},
  booktitle = {Advances in Neural Information Processing Systems},
  volume = {36},
  pages = {7093--7115},
  year = {2023},
  doi = {10.52202/075280-0310},
}

@inproceedings{dare,
  author = {Yu, Le and Yu, Bowen and Yu, Haiyang and Huang, Fei and Li, Yongbin},
  title = {{Language Models are Super Mario: Absorbing Abilities from Homologous Models as a Free Lunch}},
  booktitle = {Proc. ICML},
  volume = {235},
  pages = {57755--57775},
  year = {2024},
}

@misc{della,
  author = {Pala Tej Deep and Rishabh Bhardwaj and Soujanya Poria},
  title = {{DELLA-Merging: Reducing Interference in Model Merging through Magnitude-Based Sampling}},
  year = {2024},
  note = {arXiv:2406.11617},
}

@inproceedings{agarwal2024gkd,
  author = {Agarwal, Rishabh and Vieillard, Nino and Zhou, Yongchao and Stanczyk, Piotr and Ramos Garea, Sabela and Geist, Matthieu and Bachem, Olivier},
  title = {{On-Policy Distillation of Language Models: Learning from Self-Generated Mistakes}},
  booktitle = {Proc. ICLR},
  pages = {21246--21263},
  year = {2024},
}

@inproceedings{gu2024minillm,
  author = {Yuxian Gu and Li Dong and Furu Wei and Minlie Huang},
  title = {{Mini{LLM}: Knowledge Distillation of Large Language Models}},
  booktitle = {Proc. ICLR},
  year = {2024},
}

@inproceedings{oh2026kl,
  author = {Minjae Oh and Sangjun Song and Gyubin Choi and Yunho Choi and Yohan Jo},
  title = {{KL for a KL: On-Policy Distillation with Control Variate Baseline}},
  booktitle = {Proc. NeurIPS},
  year = {2026},
  url = {https://neurips.cc/virtual/2026/poster/149290},
}

@misc{pinckney2025cvdp,
  author = {Nathaniel Pinckney and Chenhui Deng and Chia-Tung Ho and Yun-Da Tsai and Mingjie Liu and Wenfei Zhou and Brucek Khailany and Haoxing Ren},
  title = {{Comprehensive Verilog Design Problems: A Next-Generation Benchmark Dataset for Evaluating Large Language Models and Agents on RTL Design and Verification}},
  year = {2025},
  note = {arXiv:2506.14074},
}
\end{document}